\documentclass[letterpaper]{article} 
\usepackage{aaai25}  
\usepackage{times}  
\usepackage{helvet}  
\usepackage{courier}  
\usepackage[hyphens]{url}  
\usepackage{graphicx} 
\usepackage{natbib}  
\usepackage{caption} 
\nocopyright
\usepackage{algorithm}
\usepackage{algorithmic}
\usepackage{booktabs}
\usepackage{newfloat}
\usepackage{listings}
\DeclareCaptionStyle{ruled}{labelfont=normalfont,labelsep=colon,strut=off} 
\floatstyle{ruled}
\newfloat{listing}{tb}{lst}{}
\floatname{listing}{Listing}
\title{MathSpeech: Leveraging Small LMs for Accurate Conversion in Mathematical Speech-to-Formula}
\author{
    Sieun Hyeon\textsuperscript{\rm 1}\equalcontrib,
    Kyudan Jung\textsuperscript{\rm 2}\equalcontrib,
    Jaehee Won\textsuperscript{\rm 3}, 
    Nam-Joon Kim\textsuperscript{\rm 1}\footnotemark[2], \\
    Hyun Gon Ryu\textsuperscript{\rm 4},
    Hyuk-Jae Lee\textsuperscript{\rm 1, 5},
    Jaeyoung Do\textsuperscript{\rm 1, 5}\footnote{Corresponding authors}
}
\affiliations{
    \textsuperscript{\rm 1}Department of Electrical and Computer Engineering, Seoul National University \\
    \textsuperscript{\rm 2}Department of Mathematics, Chung-Ang University \\
    \textsuperscript{\rm 3}College of Liberal Studies, Seoul National University \\
    \textsuperscript{\rm 4}NVIDIA \\
    \textsuperscript{\rm 5}Interdisciplinary Program in Artificial Intelligence, Seoul National University 

   zxc2692@snu.ac.kr , wjdrbeks1021@cau.ac.kr , wonjaehe\_e@snu.ac.kr ,
   knj01@snu.ac.kr ,   \\ hryu@nvidia.com , hjlee@capp.snu.ac.kr , jaeyoung.do@snu.ac.kr 
}

\usepackage{bibentry}

\begin{document}

\maketitle

\begin{abstract}

In various academic and professional settings, such as mathematics lectures or research presentations, it is often necessary to convey mathematical expressions orally. However, reading mathematical expressions aloud without accompanying visuals can significantly hinder comprehension, especially for those who are hearing-impaired or rely on subtitles due to language barriers. For instance, when a presenter reads Euler's Formula, current Automatic Speech Recognition (ASR) models often produce a verbose and error-prone textual description (e.g., e to the power of i x equals cosine of x plus i \textit{side} of x), instead of the concise \LaTeX{} format (i.e., $ e^{ix} = \cos(x) + i\sin(x) $), which hampers clear understanding and communication. To address this issue, we introduce MathSpeech, a novel pipeline that integrates ASR models with small Language Models (sLMs) to correct errors in mathematical expressions and accurately convert spoken expressions into structured \LaTeX{} representations. 
Evaluated on a new dataset derived from lecture recordings, MathSpeech demonstrates \LaTeX{} generation capabilities comparable to leading commercial Large Language Models (LLMs), while leveraging fine-tuned small language models of only 120M parameters.
Specifically, in terms of CER, BLEU, and ROUGE scores for \LaTeX{} translation, MathSpeech demonstrated significantly superior  capabilities compared to GPT-4o. We observed a decrease in CER from 0.390 to 0.298, and higher ROUGE/BLEU scores compared to GPT-4o.


\end{abstract}

%
\begin{links}
    \link{Code}{https://github.com/hyeonsieun/MathSpeech}
\end{links}

\section{Introduction}

Taking lectures that cover mathematical content through videos and watching academic presentations via video recordings or online streaming is no longer something out of the ordinary. Online video platforms have revolutionized mathematical education by making high-quality lectures and resources accessible to a global audience, breaking down geographical and financial barriers. However, these platforms face a critical challenge: accurately generating subtitles for mathematics lectures. Although platforms like YouTube offer automatic subtitle services, their performance deteriorates markedly when handling mathematical content, particularly equations and formulas. As a result, some providers (such as MIT OpenCourseWare\footnote{https://www.youtube.com/@mitocw}) are compelled to manually create subtitles, but this requires a tremendous amount of effort and human labeling resources.

The crux of this issue lies in the limitations of current Automated Speech Recognition (ASR) models when confronted with mathematical expressions. Despite advancements in ASR technology \cite{radford2023robust, canary, Parakeet, gandhi2023distil}, these models significantly underperform in recognizing and transcribing mathematical speech. This shortcoming became apparent during our initial investigations, highlighting the absence of a benchmark dataset specifically for evaluating ASR models' proficiency in mathematical contexts.

To address this gap, we first developed a novel benchmark dataset comprising 1,101 audio samples from real mathematics lectures available on YouTube. This dataset\footnote{https://huggingface.co/datasets/AAAI2025/MathSpeech} serves as a crucial tool for assessing the capabilities of various ASR models in mathematical speech recognition. Our evaluations using this dataset revealed not only poor performance in equation transcription but also a critical lack of \LaTeX{} generation ability, which is the standard for typesetting mathematical equations, in existing ASR models. This limitation significantly impedes learners' comprehension, especially when dealing with complex mathematical expressions. An example of this case is presented in Table \ref{table1}.

\begin{table}[h]
\centering
\small
\begin{tabular}{@{}p{1.75cm}|p{5.9cm}}
    \toprule
    \textbf{Formula}  & $ e^{ix} = \cos(x) + i\sin(x) $ \\
    \midrule
    \textbf{Spoken English(SE)} & e to the power of i x equals cosine of x plus i sine of x\\
    \midrule 
    \textbf{ASR result with error} & e to the power of i x equals cosine of x plus i \textit{side} of x\\
    \bottomrule
\end{tabular}
\caption{
Comparison between the actual equation and the ASR result}
\label{table1}
\end{table}

To address these challenges, we introduce MathSpeech, an innovative ASR pipeline specifically designed to transcribe mathematical speech directly into \LaTeX{} code instead of plain text. This approach enhances the learning experience by enabling the accurate rendering of mathematical expressions in subtitles, thereby supporting learners in their math education. Rather than incurring the significant costs associated with fine-tuning ASR models on domain-specific mathematical speech data—especially given the lack of publicly available datasets—we developed a novel method that converts mathematical speech into \LaTeX{} using small Language Models (LMs). Our methodology corrects ASR outputs containing mathematical expressions (even if they contain errors) and transforms the corrected output into \LaTeX{} using fine-tuned small LMs. By employing effective fine-tuning techniques that account for ASR errors and the nuances of spoken English in mathematical contexts, our pipeline has demonstrated superior \LaTeX{} generation capabilities compared to commercial large language models like GPT-4o \cite{gpt4o} and Gemini-Pro \cite{gemini}, despite using a relatively small language model with only 120M parameters.

In summary, our contributions are as follows:

\begin{itemize}
    \item We constructed and released the first benchmark dataset for evaluating ASR models' ability to transcribe mathematical equations.
    \item We identified and demonstrated the poor performance of existing ASR models in reading mathematical equations.
    \item We proposed a pipeline that corrects ASR errors and converts the output into \LaTeX{}
    \item We confirmed that our pipeline, despite being significantly smaller (120M parameters) than commercial LLMs, outperformed GPT-4o \cite{gpt4o} and  Gemini-Pro \cite{gemini}.
\end{itemize}

\section{Related Works}

\subsection{ASR correction with LM}

As ASR \cite{radford2023robust, canary, Parakeet, gandhi2023distil} systems have advanced, the need to correct ASR errors has become increasingly important. To enhance the quality of ASR output, there has been a significant amount of prior research focused on post-processing using language models. Since ASR outputs are in text form, many studies have employed sequence-to-sequence techniques.

In the past, statistical machine translation \cite{cucu2013statistical, dharo16_interspeech} was used for this purpose. With the development of neural network-based language models, autoregressive sequence-to-sequence models are used for error correction \cite{tanaka18_interspeech, 10.1145/3557894}, like neural machine translation.
Moreover, with the advancement of attention mechanisms \cite{7472621, 9640576}, research utilizing the Transformer architecture \cite{vaswani2017attention} for error correction has demonstrated strong performance \cite{9053126, NEURIPS2021_b597460c, leng2023softcorrect}. Additionally, research on ASR error correction has been conducted using various language models, such as BERT \cite{yang22g_interspeech}, BART \cite{zhao21_interspeech}, ELECTRA \cite{9688175, yeen23_interspeech}, and T5 \cite{ma23e_interspeech, yeen23_interspeech}.

Moreover, with the emergence of various Large Language Models (LLMs), which have shown remarkable performance across diverse domains, research on post-processing for ASR correction has also been actively conducted \cite{hu2024large, sachdev2024evolutionarypromptdesignllmbased, ma2023generativelargelanguagemodels, hu-etal-2024-listen}. Several studies have shown that LLMs can be effectively used for ASR correction. However, LLMs have certain drawbacks, such as their large size and slow inference speed.

When using LLMs for ASR error correction, some research has demonstrated that utilizing multiple candidates generated during beam search can result in a voting effect \cite{leng-etal-2021-fastcorrect-2}, leading to improved performance. This method, known as N-best, allows the N candidates obtained from the ASR output to provide clues to the language model regarding potential errors. Many studies \cite{imamura-sumita-2017-ensemble, 9688210, ganesan-etal-2021-n, ma23f_interspeech, ma23e_interspeech, leng2023softcorrect, 9053213} on ASR correction have adopted this N-best approach.

\subsection{\LaTeX{} translation and generation}

Research related to \LaTeX{} has mainly focused on converting formula images to \LaTeX{} \cite{blecher2023nougatneuralopticalunderstanding, pix2tex} using Optical Character Recognition(OCR). Recently, studies have also been conducted on generating \LaTeX{} from spoken English that describes formulas \cite{jung2024mathbridge}. However, this research has not explored correcting ASR results obtained from actual speech before translating them into \LaTeX{}. The focus of this study has been on translating clean, error-free spoken English into \LaTeX{}, without considering ASR errors

Additionally, research has been conducted on using Large Language Models (LLMs) to enhance efficiency and quality in popular academic writing tools like Overleaf \cite{wen2024overleafcopilot}, which includes generating \LaTeX{} with LLMs. In fact, well-known state-of-the-art commercial LLMs such as GPT series \cite{gpt3_5, gpt4o}, Gemini-Pro \cite{gemini} have demonstrated remarkable abilities in generating \LaTeX{}. However, the previous study \cite{jung2024mathbridge, jung2024texbleuautomaticmetricevaluate,10890531} indicates that comparing the \LaTeX{} generation and translation capabilities of different LLMs presents a significant challenge, as there is no suitable metric to measure \LaTeX{} generation performance.

\section{Motivation}
Subtitle services are often used when individuals watch academic videos or lectures. For the general public, subtitles serve as an auxiliary tool to help them understand video content. However, for individuals with hearing impairments or students who speak a different language than the lecturer, subtitles are essential. Inaccurate subtitle services can severely hinder content comprehension, leading to a significant decrease in learning effectiveness. With recent advancements in Automatic Speech Recognition (ASR) models, the ability to convert speech into text has become highly accurate, greatly benefiting these users. However, the accuracy of ASR models remains significantly lower for academic videos in fields such as mathematics and physics than for other subjects.

\begin{table}[h]
\centering
\small
\begin{tabular}{@{}p{1cm}|p{2.2cm}|p{1cm}|p{1.3cm}|p{1.2cm}@{}}
    \toprule
    & Models & Params & WER(\%) (Leaderboard) & WER(\%) (Formula)  \\
    \midrule
    OpenAI & Whisper-base & 74M & 10.3 & 34.7 \\
    & Whisper-small & 244M & 8.59 & 29.5 \\
    & Whisper-largeV2 & 1550M & 7.83 & 31.0 \\
    & Whisper-largeV3 & 1550M & 7.44 & 33.3  \\
    \midrule
    NVIDIA & Canary-1B & 1B & 6.5 & 35.2 \\
    \bottomrule
\end{tabular}
\caption{The WER for Leaderboard was from the HuggingFace Open ASR Leaderboard, while the WER for Formula was measured using our MathSpeech Benchmark.}
\label{table2}
\end{table}

Table \ref{table2} presents the results of measuring the WER (Word Error Rate) of ASR models using mathematical speech collected from actual lecture videos.

When comparing the WER results of Whisper \cite{radford2023robust} and Canary-1b \cite{canary} on the HuggingFace Open ASR Leaderboard\footnote{https://huggingface.co/spaces/hf-audio/open\_asr\_leaderboard}\footnote{This value is based on results as of 2024-08-16.} with the MathSpeech benchmark dataset WER results, we observed that the WER for formula speech was significantly higher. The reasons for the elevated error rates are as follows.

\noindent \textbf{(1) Severe Noise}: Our benchmark dataset consists of audio from lecture videos recorded 10 to 20 years ago, resulting in relatively poor audio quality and higher levels of noise. Additionally, since the audio is taken from real lectures, it includes sounds such as chalk writing on the blackboard and students' chatter mixed into the speech.

\noindent \textbf{(2) Non-native Accent}: When the speaker is not a native English speaker, the model often misinterprets some words as other words due to the speaker's accent. Our benchmark dataset includes speakers with distinctive accents.

\noindent \textbf{(3) Label Ambiguity Problem}: Mathematical expressions read aloud can be ASR-transcribed in multiple ways, which increases the WER. In other words, while the meaning is correct, the text output differs, leading it to be counted as an error. For example, in the case of speech reading number \textbf{1}, we transcribed it as \textbf{1}, but the ASR model outputs it as \textbf{one}. Although the ASR output for mathematical speech differed from the labels we assigned, it was semantically equivalent, which contributed to the higher WER measurements.

To accurately convert the spoken English of mathematical expressions into \LaTeX{}, it is necessary to carefully consider these factors and adjust the ASR errors accordingly. Therefore, we propose a method for training small language models that considers both the \LaTeX{} output and ASR transcription results. This will be discussed in the following section.

\begin{figure*}[t]
\centering
\includegraphics[width=\textwidth]{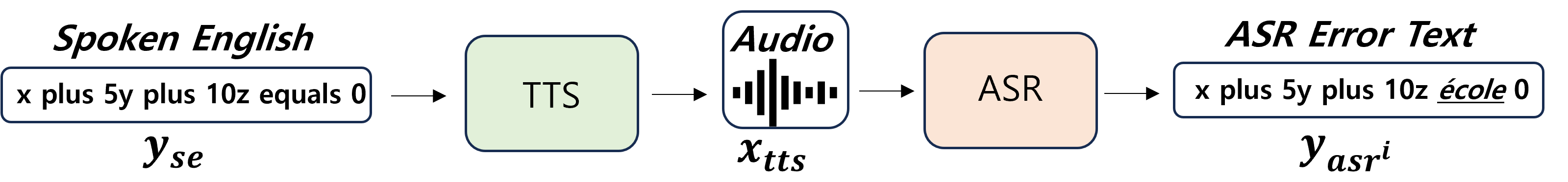} 
\caption{Method for Collecting ASR Error Results.}
\label{fig1}
\end{figure*}

\begin{figure*}[t]
\centering
\includegraphics[width=\textwidth]{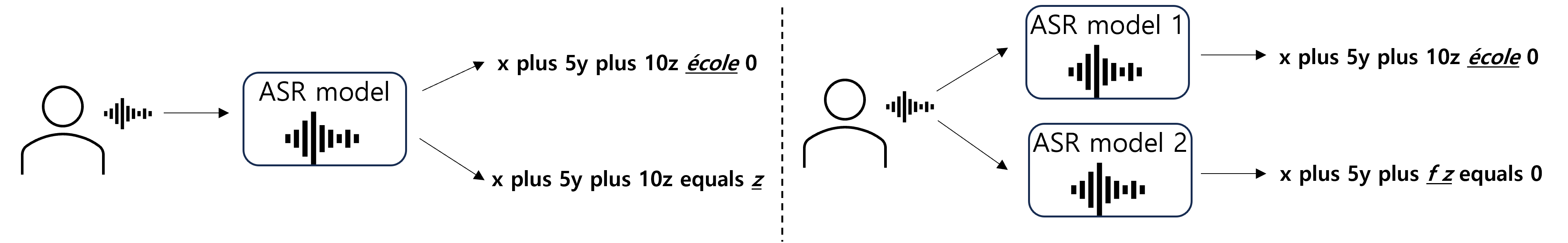} 
\caption{This figure compares 2-beam search and our method. The left shows top-2 beam search by a single ASR model, while the right shows top-1 beam search by two ASR models.}
\label{fig2}
\end{figure*}

\section{Methodology}

\subsection{Datasets}

Our goal is to implement a lightweight, fast, and high-performance pipeline that outputs \LaTeX{}. We determined that the most effective approach is fine-tuning small LMs, which involves collecting appropriate datasets. To achieve this, we collected two types of data.

\noindent \textbf{(1) (Spoken English, \LaTeX{}) pairs}

\noindent As mentioned earlier, ASR models convert spoken mathematical expressions into plain English text. Since our goal is to translate such Spoken English(SE) into \LaTeX{}, we decided to use a dataset of (Spoken English, \LaTeX{}) pairs. We were able to obtain a publicly available dataset \cite{jung2024mathbridge} on HuggingFace and used it in our work. This data was collected by web crawling and OCR, extracting only the \LaTeX{} portions from arxiv papers and textbooks. This is a large dataset containing 23 million (SE, \LaTeX{}) data pairs.
We fine-tuned the T5-small \cite{raffel2023exploringlimitstransferlearning} with this dataset to convert Spoken English(SE) to \LaTeX{}.

\noindent \textbf{(2) ASR Error Correction Dataset}

\noindent In our initial experiments, we attempted to fine-tune T5 \cite{raffel2023exploringlimitstransferlearning} only using the (Spoken English(SE), \LaTeX{}) pair dataset and connect it to the ASR as a post-LM. However, the performance was not satisfactory because the ASR model itself made significant errors when converting spoken mathematical expressions into plain text (Table 3). 

\begin{table}[h]
\centering
\small
\begin{tabular}{@{}p{3.9cm}|p{3.9cm}}
    \toprule
    \textbf{Spoken English(SE)} & \textbf{ASR error result} \\
    \midrule
    x plus 5y plus 10z equals 0 & x plus 5y plus 10z \textit{école} 0  \\
    \midrule
    cosine of psi sub i, psi sub j & \textit{Posing} of psi sub i, psi sub j \\
    \bottomrule
\end{tabular}
\caption{The examples of ASR error results}
\label{table4}
\end{table}

Since such erroneous texts did not exist in the dataset \cite{jung2024mathbridge}, the fine-tuned T5 also produced incorrect \LaTeX{} outputs. Therefore, we determined that adding a process to correct errors that occur in the ASR system would significantly improve performance. Thus, we built another dataset for fine-tuning the ASR error correction model. As shown in Figure 1, we converted Spoken English (SE) into speech using TTS and then fed the speech into the ASR model to obtain the erroneous Spoken English ASR outputs.

In other words, let $y_{se}$ represent Spoken English and $y_{latex}$ represent the \LaTeX{} expression for $y_{se}$. Then, we can denote the data pair \cite{jung2024mathbridge} as $(y_{se}$, $y_{latex})$.

Using TTS, we generate audio for $y_{se}$. If we denote this audio as $x_{tts}$, then we can denote the text obtained by inputting $x_{tts}$ into an ASR model as $y_{asr^i}$, where $i$ is used to distinguish between different models. Therefore, using the method shown in Figure 1, we transform $(y_{se}$, $y_{latex})$ into $(y_{asr^i}$ , $y_{se}$ , $y_{latex})$.

At this stage, we used the VITS \cite{pmlr-v139-kim21f} model for TTS and collected the voices using 2-speaker. For the ASR model, we used Whisper-base, small, largeV2 \cite{radford2023robust} and Canary-1b \cite{canary} to collect the Error ASR results. We collected 6M ASR error results using Whisper-base and small, and 1M ASR error results using Whisper-largeV2 and Canary-1b.

\subsection{Models}

Our MathSpeech pipeline can be seen in Figure 3. We used two models configured in two stages.

\begin{figure*}[t]
\centering
\includegraphics[width=\textwidth]{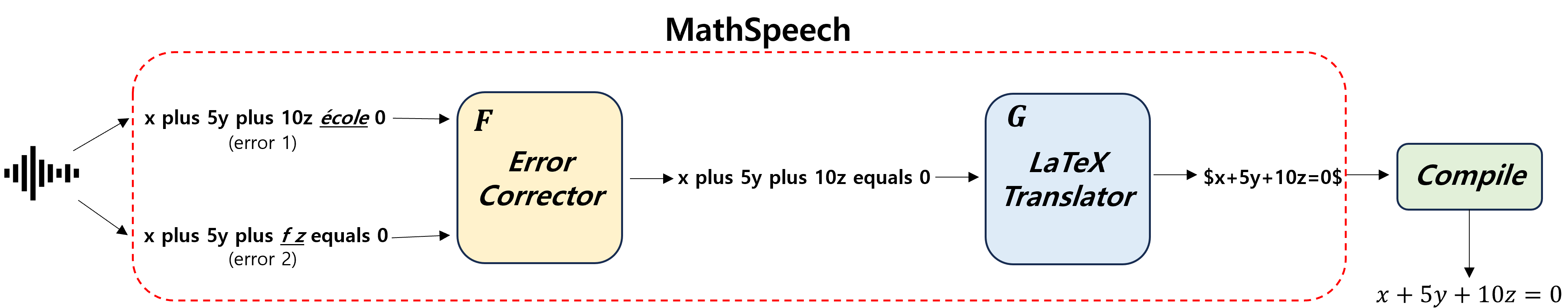} 
\caption{Our pipeline that converts the lecturer's voice into \LaTeX{}.}
\label{fig3}
\end{figure*}

\begin{figure*}[t]
\centering
\includegraphics[width=\textwidth]{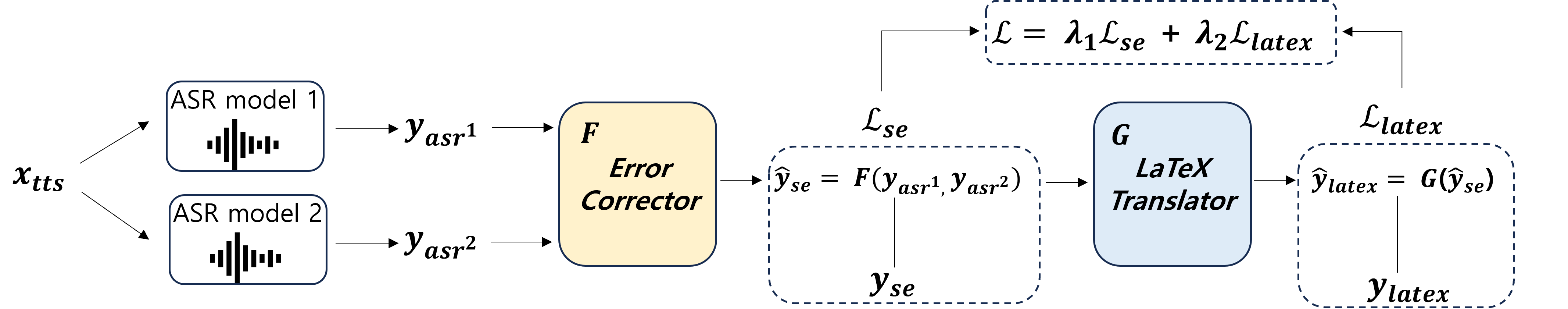} 
\caption{The method of training MathSpeech in an end-to-end manner}
\label{fig4}
\end{figure*}

\noindent \textbf{(1) Error Corrector}

\noindent The purpose of the Error Corrector is to fix the errors that occur in the ASR model. In other words, the goal of the Error Corrector is to make the ASR error results similar to the input data of the \LaTeX{} Translator, which is called Spoken English. In previous studies, T5 \cite{ma23e_interspeech, ma23f_interspeech, yeen23_interspeech} has demonstrated good performance in ASR error correction. To minimize the model size, we used T5-small for the Error Corrector.

Additionally, recent research has shown that inputting multiple candidates \cite{imamura-sumita-2017-ensemble, 9688210, ganesan-etal-2021-n, ma23f_interspeech, ma23e_interspeech, leng2023softcorrect, 9053213} generated by ASR beam search can yield good performance in various ASR Corrector models due to the voting effect \cite{leng-etal-2021-fastcorrect-2}. We have further developed this idea. Instead of using multiple candidates generated by ASR beam search, we use multiple candidates from the top-1 results of different models (Figure 2).

An advantage of this method is its versatility. We believe that training the Error Corrector with error information from various models will make it universally applicable to different ASR models. For example, if the corrector is trained with two ASR beam search results from Whisper-small, it may perform well on Whisper-small, but not on Whisper-base. However, if the corrector is trained with the ASR results from both Whisper-small and Whisper-base, we believe that it will effectively correct the errors for both models. Experimental results are presented in the following section.

\noindent \textbf{(2) \LaTeX{} Translator}

\noindent According to a previous study \cite{jung2024mathbridge}, the T5 model achieved the best performance for the \LaTeX{} translator. To minimize the model size, we used the smallest version, T5-small. We fine-tuned T5-small using the (SE, \LaTeX{}) pair dataset so that when Spoken English is input, \LaTeX{} is output.

\subsection{Training}

We implemented a pipeline connecting two T5-small models. Instead of simply chaining the fine-tuned Error Corrector and \LaTeX{} Translator, we performed end-to-end training (Figure 4). Considering the characteristics of ASR error results and \LaTeX{}, we constructed the loss function as follows.

For a given input audio \(x_{tts}\), let the inference results of two ASR models be \(y_{asr^1}\) and \(y_{asr^2}\). Here, \(y_{asr^1}\) and \(y_{asr^2}\) are the outputs from the ASR models, which may contain errors. We then provide these two ASR outputs to the Error Corrector \(F\). The resulting text can be denoted as \(F(y_{asr^1}, y_{asr^2})\). Next, we take \(\hat{y}_{se} = F(y_{asr^1}, y_{asr^2})\) and feed it into the \LaTeX{} Translator \(G\) to produce the corresponding \LaTeX{}, denoted as \(\hat{y}_{latex} = G(F(y_{asr^1}, y_{asr^2}))\). The loss function is then defined as follows:

\begin{equation}
    \mathcal{L} = \lambda_{1}\mathcal{L}_{se} + \lambda_{2}\mathcal{L}_{latex}
\end{equation}

And $\mathcal{L}_{se}$ and $\mathcal{L}_{latex}$ are calculated as cross-entropy losses of the tokenized outputs $\hat{y}_{se}$ and ${y}_{se}$, and $\hat{y}_{latex}$ and ${y}_{latex}$, respectively.

\begin{equation}
    \mathcal{L}_{se} = -\sum_{t=1}^{T} \sum_{i=1}^{V} y_{se_t^i} \log(\hat{y}_{se_t^i})
\end{equation}

\begin{equation}
    \mathcal{L}_{latex} = -\sum_{t=1}^{T} \sum_{i=1}^{V} y_{latex_t^i} \log(\hat{y}_{latex_t^i})
\end{equation}

The cross-entropy loss is calculated as a negative sum over all time steps \(t\) from 1 to \(T\) (the length of the sequence) and over all possible words \(i\) in the vocabulary \(V\). At each time step \(t\), \(y_{se_t}\) is a one-hot encoded vector over the vocabulary, where \(y_{se_t^i}\) is its \(i\)-th element. Specifically, \(y_{se_t^i} = 1\) if the correct ground-truth word at time step \(t\) is the \(i\)-th word in the vocabulary, and \(0\) otherwise. Correspondingly, \(\hat{y}_{se_t}\) is the predicted probability distribution vector over the vocabulary at time \(t\), and \(\hat{y}_{se_t^i}\) is the predicted probability that the correct word at time \(t\) is the \(i\)-th word in the vocabulary.

We calculated the final loss $\mathcal{L}$ by assigning different weights to the two cross entropy loss functions. The weight assigned to the $\hat{y}_{latex}$ was set higher than the weight for the loss on $\hat{y}_{se}$ because there can be different SE for the same \LaTeX{} result (Table 4).

\begin{table}[h]
\centering
\small
\begin{tabular}{@{}p{1.2cm}|p{6.4cm}}
    \toprule
    \textbf{1st SE} & x plus five y plus ten z equals zero\\ 
    \textbf{2nd SE} & x plus 5 y plus 10 z equals 0  \\
    \midrule
    \LaTeX{}  & \$x + 5y + 10z = 10\$ \\
    \bottomrule
    
\end{tabular}
\caption{The reason why errors between \LaTeX{} are more critical than errors between SE.}
\label{table4}
\end{table}

In other words, even if the SEs are slightly different, the \LaTeX{} can still produce the correct answer because they are semantically equivalent. Therefore, our goal is to obtain the correct \LaTeX{}, so we assigned a higher weight to $\mathcal{L}_{latex}$ when constructing the loss function. Our experimental results showed that setting the weight $\lambda_{1}$ for SE to 0.3 and the weight for \LaTeX{} to 0.7 yielded the best performance, so we used these values.

\begin{table*}[t]
\centering
\small
\begin{tabular}{p{2.4cm}|p{2.2cm}|p{1.8cm}|p{1.9cm}|p{1.9cm}|p{1.6cm}} 
\toprule
\textbf{ASR} &
\textbf{LM} &
\textbf{CER} $\downarrow$ &
\textbf{ROUGE-1} $\uparrow$ &
\textbf{ROUGE-L} $\uparrow$ &
\textbf{BLEU} $\uparrow$ \\
\midrule
whisper-base &
GPT-3.5 &
0.443 &
0.782 &
0.775 &
0.483 \\

 &
GPT-4o &
\textit{0.410}$^{*}$ &
\textit{0.813}$^{*}$ &
\textit{0.808}$^{*}$ &
\textit{0.487}$^{*}$ \\

 &
Gemini-Pro &
0.424 &
0.756 &
0.749 &
0.418 \\

 &
\textbf{MathSpeech} &
\textbf{0.336} &
\textbf{0.824} &
\textbf{0.819} &
\textbf{0.662} \\

\midrule
whisper-small &
GPT-3.5 &
0.391 &
0.815 &
0.809 &
\textit{0.519}$^{*}$ \\

 &
GPT-4o &
\textit{0.384}$^{*}$ &
\textit{0.840}$^{*}$ &
\textit{0.835}$^{*}$ &
0.516 \\

 &
Gemini-Pro &
0.390 &
0.799 &
0.792 &
0.455 \\

 &
\textbf{MathSpeech} &
\textbf{0.309} &
\textbf{0.852}$^{++}$ &
\textbf{0.847}$^{++}$ &
\textbf{0.689}$^{++}$ \\
\midrule
whisper-largeV2 &
GPT-3.5 &
0.401 &
0.820 &
0.813 &
0.519 \\

 &
GPT-4o &
\textit{0.380}$^{*}$ &
\textit{0.844}$^{*}$ &
\textit{0.839}$^{*}$ &
\textit{0.520}$^{*}$ \\

 &
Gemini-Pro &
0.393 &
0.800 &
0.792 &
0.458 \\

 &
\textbf{MathSpeech} &
\textbf{0.298}$^{++}$ &
\textbf{0.848} &
\textbf{0.844} &
\textbf{0.683} \\
\midrule
whisper-largeV3 &
GPT-3.5 &
0.404 &
0.797 &
0.792 &
0.493 \\

 &
GPT-4o &
\textit{0.398}$^{*}$ &
\textbf{0.826} &
\textbf{0.822} &
\textit{0.495}$^{*}$ \\

 &
Gemini-Pro &
0.409 &
0.772 &
0.767 &
0.431 \\

 &
\textbf{MathSpeech} &
\textbf{0.317} &
\textit{0.817}$^{*}$ &
\textit{0.812}$^{*}$ &
\textbf{0.673} \\
\midrule
canary-1b &
GPT-3.5 &
0.458 &
0.795 &
0.787 &
0.434 \\

 &
GPT-4o &
\textit{0.422}$^{*}$ &
\textit{0.824}$^{*}$ &
\textit{0.819}$^{*}$ &
\textit{0.445}$^{*}$ \\

 &
Gemini-Pro &
0.442&
0.781 &
0.774 &
0.392 \\

 &
\textbf{MathSpeech} &
\textbf{0.325} &
\textbf{0.832} &
\textbf{0.828} &
\textbf{0.674} \\

\bottomrule
\end{tabular}
\caption{Having only one model in the ASR column means that the top-2 beam search ASR outputs of a single model are input into the LLM. The bold text indicates the result with the best score for the same ASR output, while an asterisk ($*$) indicates the second-best score for the same ASR output. The double plus sign (++) indicates the case with the best score.}
\label{table5}
\end{table*}

\begin{table*}[t]
\centering
\small
\begin{tabular}{p{2.4cm}|p{2.2cm}|p{1.8cm}|p{1.9cm}|p{1.9cm}|p{1.6cm}} 
\toprule
\textbf{ASR} &
\textbf{LM} &
\textbf{CER} $\downarrow$ &
\textbf{ROUGE-1} $\uparrow$ &
\textbf{ROUGE-L} $\uparrow$ &
\textbf{BLEU} $\uparrow$ \\
\midrule

whisper-base &
GPT-3.5 &
0.407 &
0.801 &
0.792 &
0.502 \\

\& &
GPT-4o &
\textit{0.379}$^{*}$ &
\textit{0.843}$^{*}$ &
\textit{0.839}$^{*}$ &
\textit{0.518}$^{*}$ \\

whisper-small &
Gemini-Pro &
0.382 &
0.802 &
0.796 &
0.457 \\

 &
\textbf{MathSpeech} &
\textbf{0.243}$^{++}$ &
\textbf{0.870}$^{++}$ &
\textbf{0.864}$^{++}$ &
\textbf{0.718}$^{++}$ \\

\midrule
whisper-small &
GPT-3.5 &
0.388 &
0.825 &
0.818 &
0.523 \\

\& &
GPT-4o &
\textit{0.373}$^{*}$ &
\textit{0.852}$^{*}$ &
\textit{0.848}$^{*}$ &
\textit{0.532}$^{*}$ \\

whisper-largeV2 &
Gemini-Pro &
0.374 &
0.806 &
0.800 &
0.469 \\

 &
\textbf{MathSpeech} &
\textbf{0.269} &
\textbf{0.864} &
\textbf{0.859} &
\textbf{0.708} \\

\midrule
canary-1b &
GPT-3.5 &
0.399 &
0.826 &
0.820 &
\textit{0.524}$^{*}$ \\

\& &
GPT-4o &
\textit{0.386}$^{*}$ &
\textbf{0.849} &
\textbf{0.844} &
0.516 \\

whisper-largeV2 &
Gemini-Pro &
0.393 &
0.800 &
0.793 &
0.458 \\

 &
\textbf{MathSpeech} &
\textbf{0.294} &
\textit{0.848}$^{*}$ &
\textit{0.843}$^{*}$ &
\textbf{0.694} \\

\midrule

canary-1b &
GPT-3.5 &
0.394 &
0.813 &
0.806 &
0.504 \\

\& &
GPT-4o &
\textit{0.375}$^{*}$ &
\textit{0.849}$^{*}$ &
\textit{0.844}$^{*}$ &
0.514 \\

whisper-largeV3 &
Gemini-Pro &
0.399 &
0.810 &
0.804 &
0.445 \\

 &
\textbf{MathSpeech} &
\textbf{0.292} &
\textbf{0.853} &
\textbf{0.846} &
\textbf{0.698} \\
\hline
\hline
\textit{Total Average} &
GPT-3.5 &
0.409 &
0.808 &
0.801 &
0.500 \\

 &
GPT-4o &
\textit{0.390}$^{*}$ &
\textit{0.838}$^{*}$ &
\textit{0.833}$^{*}$ &
\textit{0.505}$^{*}$ \\

 &
Gemini-Pro &
0.400 &
0.792 &
0.785 &
0.442 \\

 &
\textbf{MathSpeech} &
\textbf{0.298} &
\textbf{0.845} &
\textbf{0.840} &
\textbf{0.689} \\
\bottomrule
\end{tabular}
\caption{Having two models in the ASR column means that the top-1 beam search ASR output from each of the two different models is obtained and both are input into the LLM. \textit{Total Average} refers to the average of all the results from Table 5 and 6.}
\label{table6}
\end{table*}

\begin{table*}[t]
\centering
\small
\begin{tabular}{p{2cm}|p{5.2cm}|p{1.4cm}|p{1.9cm}|p{1.9cm}|p{1.6cm}} 
\toprule
\textbf{ASR} &
\textbf{LM} &
\textbf{CER} $\downarrow$ &
\textbf{ROUGE-1} $\uparrow$ &
\textbf{ROUGE-L} $\uparrow$ &
\textbf{BLEU} $\uparrow$ \\
\midrule
whisper-base &
T5-small (1 stage, w/o corrector) &
0.693 &
0.734 &
0.729 &
0.483 \\

&
T5-base (1 stage, w/o corrector) &
\textit{0.530}$^{*}$ &
\textit{0.749}$^{*}$ &
\textit{0.744}$^{*}$ &
\textit{0.554}$^{*}$ \\

&
\textbf{MathSpeech} &
\textbf{0.336} &
\textbf{0.824} &
\textbf{0.819} &
\textbf{0.662} \\

\midrule
whisper-small &
T5-small (1 stage, w/o corrector) &
0.630 &
0.769 &
0.765 &
0.521 \\

&
T5-base (1 stage, w/o corrector) &
\textit{0.423}$^{*}$ &
\textit{0.783}$^{*}$ &
\textit{0.780}$^{*}$ &
\textit{0.618}$^{*}$ \\

&
\textbf{MathSpeech} &
\textbf{0.309} &
\textbf{0.852} &
\textbf{0.847} &
\textbf{0.689} \\

\midrule
whisper-base &
T5-small (1 stage, fine-tuned with errors)&
0.403 &
\textit{0.824}$^{*}$ &
\textit{0.819}$^{*}$ &
0.635 \\

\& &
T5-base (1 stage, fine-tuned with errors) &
0.358 &
0.813 &
0.809 &
0.650 \\

whisper-small &
T5-small (2 stage, Just connect)&
0.357 &
0.820 &
0.815 &
\textit{0.656}$^{*}$ \\
 &
T5-base (2 stage, Just connect)&
\textit{0.343}$^{*}$ &
0.817 &
0.814 &
0.651 \\
&
\textbf{MathSpeech} &
\textbf{0.243} &
\textbf{0.870} &
\textbf{0.864} &
\textbf{0.718} \\

\bottomrule
\end{tabular}
\caption{Ablation Study Results. 1 stage refers to using a single T5 model, while 2 stage refers to using two T5 models.}
\label{table7}
\end{table*}

\subsection{Evaluation Metrics}

In previous studies \cite{jung2024mathbridge}, when evaluating \LaTeX{}, metrics commonly used in translation tasks, such as ROUGE \cite{lin-2004-rouge} and BLEU \cite{Papineni02bleu:a}, were employed. Based on this idea, we used ROUGE-1, BLEU, and ROUGE-L. Furthermore, we employed CER, a traditional ASR metric. However, WER was not measured in our experiments. This is because the spaces are often ignored in \LaTeX{}. For example, \$A B\$ and \$AB\$ result in the same \LaTeX{} compilation output. So the evaluation was conducted after removing all spaces.

\section{Experiments}

To evaluate the \LaTeX{} translation capabilities of our pipeline, we compared its performance against existing commercial large language models.

\subsection{Setup}
We fixed the same hyperparameters for the fine-tuned models. The maximum number of training epochs was set to 20, and the model with the lowest validation loss was selected. The learning rate was set to a maximum of 1e-4 and a minimum of 1e-6, adjusted using a linear learning rate scheduler. For the Error Corrector, which requires two ASR outputs as input, the maximum input sequence length was set to 540, with an output length of 275. For the \LaTeX{} translator, both input and output sequence lengths were set to 275. T5-small was trained with a batch size of 48 on an NVIDIA A100, and T5-base with a batch size of 84 on an NVIDIA H100.

As a comparison group for our pipeline, we selected GPT-3.5 \cite{gpt3_5}, GPT-4o \cite{gpt4o}, and Gemini-Pro \cite{gemini}, using 1-shot prompting with one example for all. To observe \LaTeX{} translation results across various ASR models, we used five ASR models.

\subsection{Result}
The experimental results are presented in Tables 5 and 6, respectively. Table 5 lists the outcomes of inputting two ASR prediction candidates that require correction into the LM. In this regard, MathSpeech achieved the best scores for the CER, ROUGE-1, ROUGE-L, and BLEU. The LLM with the second-best performance was GPT-4o, while GPT-3.5 and Gemini-Pro showed similar results.

The main factor that lowered the performance scores of the commercial large language models was hallucination. When the ASR results were unusual or ambiguous, the baseline LLMs would either output completely different \LaTeX{} formulas or produce non-\LaTeX{} texts (e.g., Sorry, I can't understand). Additionally, whisper-small and whisper-largeV2 outperformed whisper-base, whisper-largeV3, and canary-1b in overall metric scores. This result aligns with the WER measurements of Spoken English for whisper-small and whisper-largeV2, which were relatively better, as observed in Table 2. In other words, better ASR results lead to improved \LaTeX{} translation performance. 

Table 6 lists the outcomes when the top-1 results from different ASR models were used as inputs. The highest performance scores were achieved when the ASR results from Whisper-base \& small were used as inputs. This can be attributed to the fact that our ASR error results dataset contains relatively more information from Whisper-base \& small. The second highest performance was observed when using the ASR results from Whisper-small \& largev2, which can be attributed to the relatively lower WER of these two models on our benchmark dataset. Furthermore, the strong performance on Whisper-largeV3, for which we did not collect ASR error results, demonstrates that MathSpeech can perform well even on ASR models that were not used during training. Moreover, our MathSpeech model, being a small-sized model with 120M parameters, has low latency. When inference latency was measured on an NVIDIA V100 GPU, it took 0.45 seconds to convert the ASR result of 5 seconds of speech into \LaTeX{}.

\section{Ablation Study}
To demonstrate the effectiveness of the MathSpeech structure, we conducted ablation studies (Table 7).

To show that correcting ASR outputs is crucial for \LaTeX{} translation, we removed the corrector and conducted an experiment in which ASR outputs were directly translated into \LaTeX{}. As a result, both T5-small and T5-base showed significant performance degradation.

To demonstrate that implementing the corrector and translator in a 2-stage structure is effective, we trained a single T5 model to perform both correction and translation in a 1-stage process and observed its performance. The experiment, where we trained the model to translate two ASR error outputs into \LaTeX{} with the same setup, showed lower performance compared to our pipeline.

To validate the effectiveness of our end-to-end training method, we compared it with a method where the corrector and translator were trained separately and simply concatenated. The results confirm that the proposed training method is more effective.

Since a single T5-base (220M) is larger than two T5-small models (120M), we did not apply our end-to-end training method to the 2-stage T5-base pipeline. However, we can infer that if we were to apply our training method to the simple concatenation of two T5-base models, the performance could potentially improve further.

\section{Future Works}

\noindent \textbf{(1) Defining a Metric to Solve the Label Ambiguity}

\noindent Since \LaTeX{} can represent the same formula in multiple ways, it is necessary to consider various possible cases when evaluating the performance of \LaTeX{} translation. \LaTeX{} is closer to a computer language, like SQL, than to natural language, so metrics like BLEU or CER are not perfect for evaluating \LaTeX{}. Therefore, it is necessary to implement a metric that is more suitable for evaluating \LaTeX{}.

\noindent \textbf{(2) Formula Detection in Practice and \LaTeX{} Conversion}

\noindent This research focuses only on the ability of ASR models and LMs to generate \LaTeX{}. However, to apply this to actual subtitle services, it is essential to develop the ability to detect and separate formulaic parts from speech. This can likely be achieved by training the LM not only on \LaTeX{} but also on mixed general text. Additionally, in real-world situations, the speaker may not finish verbally expressing the entire formula. It must be possible to complete such interrupted formulas into full formulas. This could be implemented through the inference capabilities of large language models (LLMs).

\section{Conclusion}

In this paper, we confirmed through a self-constructed benchmark dataset that existing ASR models lack the ability to read mathematical formulas and are unable to generate \LaTeX{}. To address this, we propose MathSpeech, a pipeline that connects ASR models with small LMs to generate \LaTeX{}. By effectively connecting two T5-small models and training them end-to-end, our approach demonstrated superior \LaTeX{} translation capabilities compared to existing commercial large language models. Our research opens up the possibility of more accurate subtitles in the field of math.

\section{Acknowledgments}

This work was supported in part by National Research Foundation of Korea (NRF) grant (RS-2024-00414981), Institute of Information \& communications Technology Planning \& Evaluation (IITP) grant (RS-2024-00399936, RS-2021-II211343, RS-2024-00454666, IITP-2024-RS-2024-00397085, IITP-2024-RS-2024-00441407), and Research Grant (0418-20240053) from Seoul National University. Samsung Memory Research Center (SMRC) provided research facilities for this work. J. Do is with ASRI, Seoul National University.

\bibliography{aaai25}

\newpage

\section{Appendix}

\subsection{The datasets used in this study}

This section provides detailed information about the datasets used to train and test MathSpeech.

\noindent \textbf{(1) Test dataset}

In our experiments, we used the MathSpeech benchmark dataset, which we created ourselves by extracting math lecture audio directly collected from MIT OpenCourseWare. The detailed information about the MathSpeech benchmark dataset is as follows.

\begin{table}[h]
\centering
\small
\begin{tabular}{@{}l|l@{}}
    \toprule
    \textbf{Property} & \textbf{Value} \\
    \midrule
    The number of files & 1,101 \\
    Total Duration & 5,583 seconds \\
    Average Duration per file & 5.07 seconds \\
    The number of speakers & 10 \\
    The number of men & 8 \\
    The number of women & 2 \\
    Source & https://www.youtube.com/@mitocw  \\
    \bottomrule
\end{tabular}
\caption{Statistics of the MathSpeech benchmark dataset.}
\label{table:mathspeech}
\end{table}

You can listen to the actual test data audio and view the paired transcriptions by accessing the link below:  
https://huggingface.co/datasets/AAAI2025/MathSpeech

\vspace{0.25cm}

\noindent \textbf{(2) Training dataset}

The detailed information about the training datasets used to implement the MathSpeech pipeline is as follows: 

\begin{itemize}
    \item For Error Corrector
\end{itemize}

\begin{table}[h]
\centering
\small
\begin{tabular}{@{}l|l|l@{}}
    \toprule
    \textbf{ASR Model} & \textbf{ASR Model Params} & \textbf{Data Count} \\
    \midrule
    Whisper-base & 74M & 6M \\
    Whisper-small & 244M & 6M \\
    Whisper-largeV2 & 1550M & 1M \\
    Canary-1b & 1B & 1M \\
    \bottomrule
\end{tabular}
\caption{Statistics of the ASR Error Correction Dataset}
\label{table:asr_models}
\end{table}

\begin{itemize}
    \item For LaTeX Translator
\end{itemize}

\begin{table}[h]
\centering
\small
\begin{tabular}{@{}l|l}
    \toprule
    \textbf{Spoken English(SE)} & \textbf{LaTeX formula} \\
    \midrule
     Phi sub H sub k of y equals E y & \texttt{\$ \textbackslash phi\_\{H\_k\} (y)=E y \$}  \\
    \midrule
     x sub zero equals x of t & \texttt{\$ x\_0=x(t) \$} \\
    \bottomrule
\end{tabular}
\caption{Examples of MathBridge dataset \cite{jung2024mathbridge}, which contains 23M pairs of (Spoken English, \LaTeX{})}
\label{table10}
\end{table}

\newpage

\subsection{Latency comparison between models}

This section presents a latency comparison between MathSpeech and the baseline LLMs. Due to its compact size of 120M, our model achieves fast inference speeds even with relatively limited GPU resources. The table below demonstrates the efficiency of MathSpeech and highlights its potential for use in subtitle generation services.

\begin{table}[h]
\centering
\small
\begin{tabular}{@{}l|l}
    \toprule
    \textit{ASR models} & \textit{Latency (sec)} \\
    \midrule
    whisper-base & 0.68 \\
    whisper-small & 0.79 \\
    \midrule
    \midrule
    \textit{Language Models} & \textit{Latency (sec)} \\
    \midrule
     \textbf{MathSpeech} & 0.45  \\
     GPT-4o & 0.65 \\
     GPT-3.5 & 0.40 \\
     gemini-pro & 1.53 \\     
    \bottomrule
\end{tabular}
\caption{Latency comparison. This experiment measures the time required to convert a single 5-second audio clip into LaTeX. The tests were conducted on a computer equipped with a single NVIDIA V100 GPU and an Intel(R) Xeon(R) Platinum 8480+ CPU. The GPU was exclusively used for MathSpeech, while GPT-4, GPT-3.5, and Gemini-Pro were accessed via API calls.}
\label{table11}
\end{table}

Since GPT and Gemini are closed-source models, their exact computing environments could not be confirmed, making it impossible to compare latency and speed under identical conditions. However, it is clear that these models run on a significant number of GPUs, demonstrating that our model achieves favorable latency even with minimal GPU resources.

\end{document}